# A Novel Hybrid Crossover based Artificial Bee Colony Algorithm for Optimization Problem

Sandeep Kumar
Jagannath University
Chaksu, Jaipur, Rajasthan,
India - 303901

Vivek Kumar Sharma,
Ph.D
Jagannath University
Chaksu, Jaipur, Rajasthan,
India - 303901

Rajani Kumari
Jagannath University
Chaksu, Jaipur, Rajasthan,
India - 303901

## ABSTRACT

Artificial bee colony (ABC) algorithm has proved its importance in solving a number of problems including engineering optimization problems. ABC algorithm is one of the most popular and youngest member of the family of population based nature inspired meta-heuristic swarm intelligence method. ABC has been proved its superiority over some other Nature Inspired Algorithms (NIA) when applied for both benchmark functions and real world problems. The performance of search process of ABC depends on a random value which tries to balance exploration and exploitation phase. In order to increase the performance it is required to balance the exploration of search space and exploitation of optimal solution of the ABC. This paper outlines a new hybrid of ABC algorithm with Genetic Algorithm. The proposed method integrates crossover operation from Genetic Algorithm (GA) with original ABC algorithm. The proposed method is named as Crossover based ABC (CbABC). The CbABC strengthens the exploitation phase of ABC as crossover enhances exploration of search space. The CbABC tested over four standard benchmark functions and a popular continuous optimization problem.

## 1. INTRODUCTION

Nature inspired meta-heuristics has become a looming and interesting field of research among researchers who are working on optimization problems. Almost all meta-heuristics make use of both randomization and local search. Due to randomization it can move away from local search to global search space. That's why meta-heuristics are best suitable for global optimization. Meta heuristic algorithms have two major components: diversification and intensification. Diversification explores the large search space and ensures that solution does not trap in local optima while intensification concentrates on best solution for convergence to optimality [1]. Population based meta-heuristics do not guarantee the optimal solution but they provide near-optimal solution for most difficult optimization problems. Researchers have analyzed such behaviors and designed algorithms that can be used to solve nonlinear and discrete optimization problems. Previous research [2, 3, 4, and 5] has shown that methods based on swarm intelligence have great power to find solutions of real world optimization problems. The algorithms that have emerged in recent years include ant colony optimization (ACO) [2], particle swarm optimization (PSO) [3], bacterial foraging optimization (BFO) [6] etc. Artificial bee colony (ABC) optimization algorithm introduced by D. Karaboga [7] is a new entry in class of swarm intelligence. This algorithm is inspired by the social behavior of honey bees when searching for quality food source. Similar to any other population based optimization algorithm, ABC consists of a population of inherent solutions. The inherent solutions are food sources of honey bees. The fitness is decided in terms of the quality of the food source that is nectar amount. ABC is relatively a simple, fast and population based stochastic search technique in the field of nature inspired algorithms.

There are two fundamental processes which drive the swarm to update in ABC: the deviation process, which enables exploring different fields of the search space, and the selection process, which ensures the exploitation of the previous experience. However, it has been shown that the ABC may occasionally stop moving toward the global optimum even though the population has not encounter to a local optimum [8]. It can be observed that the solution search equation of ABC algorithm is good at exploration but poor at exploitation [9]. Therefore, to maintain the proper balance between exploration and exploitation behavior of ABC, it is highly expected to develop a local search approach in the basic ABC to intensify the search region.

Now a day a number of researchers are moving towards ABC algorithm from all over the world. A lot of modifications are done in ABC in recent years after its inception by D. Karaboga [7]. In [10], D. Karaboga appears with an enhanced version of ABC for constrained optimization problems. He applied it in neural network training [11], to classification of medical pattern and clustering problems [12] and to solve TSP problems [13]. Artificial Bee Colony algorithm applied in Distributed Environments by A. Banharnsakun et al. [14]. Harish Sharma et al. developed DSABC [15], Balanced ABC [16], MeABC [17], and LFABC [18]. S. Pandey et al. developed a hybrid of ABC using crossover operation and applied it on TSP [19]. A survey of ABC carried out by H. Sharma et al. [20] shows its popularity in last decade.

R. Ramanathan used ABC algorithm in image compression [21]. A. Ahmad et al.[22] employed artificial bee colony (ABC) algorithm to formulate uniformly excited, non-uniformly spaced, symmetrical linear arrays. ABC algorithm is used to minimize the side lobe level (SLL) of uniformly excited linear antenna arrays by considering the element spacing as the optimization parameters. Alvarado Iniesta [23] makes use of ABC algorithm to optimize the flow of material in a manufacturing plant. They examine how to optimize the time and effort required to supply raw material to separate production lines in a manufacturing plant by minimizing the distance an operator must travel to distribute material from a warehouse to a set of different production lines with corresponding demand. A Ozen [24] developed a new artificial bee colony algorithm based modulation recognition technique for multipath fading channels.

The organization of this paper as follow: Section 2 gives brief idea about original ABC, analogy between behavior of honey bees and artificial bee colony algorithm. Section 3 introduces linear crossover operator. Section 4 outlines a new hybrid algorithm that integrates ABC and GA. Next section shows what are experimental setup and results for 4 benchmark functions. A real world problem taken in section 6 and proposed algorithm tested for it. Conclusion of this paper presented in section 7 followed by appendix with figures and references.

## 2. ARTIFICIAL BEE COLONY ALGORITHM

Artificial Bee Colony Algorithm is motivated from the intelligent food foraging behavior of honey bee insects. Honey bee swarm is one of the most intelligent swarms exists in nature; which follows collective intelligent method, while searching the food. The honey bee swarm has many qualities like bees can communicate the information, can memorize the environment, can store and share the information and take decisions based on that. According to changes in the environment, the swarm updates itself, assigns the tasks dynamically and moves further by social learning and teaching. This intelligent behavior of bees motivates researchers to simulate above foraging behavior of the bee swarm.

### 2.1 Analogy of Artificial Bee Colony Algorithm

The original model proposed by D. Karaboga [7] is composed of three major elements: employed and unemployed foragers, and food sources. The employed bees are ally with an appropriate food source. Employed bees have intimate knowledge about food source. Exploitation of food sources done by employed bees. When a food source abandoned employed bee become unemployed. The unemployed foragers are bees having no information about food sources and searching for a food source to exploit it. One can classify unemployed bees in two categories: scout bees and onlooker bees. Scout bees search at random for new food sources surrounding the hive. Onlooker bees observe the waggle dance in hive, to select a food source for exploitation. The third element is the rich food sources close to their hive. Comparatively in the optimization context, the number of food sources (that is the employed or onlooker bees) in ABC algorithm, is equivalent to the number of solutions in the population. Moreover, the location of a food source represents the position of a favorable solution to the optimization problem, since the trait of nectar of a food source represents the fitness cost (quality) of the correlated solution.

### 2.2 Phases of Artificial Bee Colony Algorithm

The search process of ABC follow three major steps [7]:

- Send the employed bees to a food source and calculate the nectar quality;
- Onlooker bees select the food sources after gathering information from employed bees and calculating the nectar quality;
- Determine the scout bees and employ them onto possible food sources.

The location of the food sources are arbitrarily selected by the bees at the initial stage and their nectar quality are measured. The employed bees then share the nectar information of the sources with the onlooker bees waiting at the dance area within the hive. After sharing this information, each employed bee returns to the food source checked during the previous cycle, as the location of the food source had been recalled and then selects new food source using its observed information in the neighborhood of the present food source. At the last stage, an onlooker bee uses the information retrieved from the employed bees at the dance area to select a good food source. The possibility for the food sources to be elected boosts with boost in its quality of nectar. Hence, the employed bee with information of a food source with the highest quality of nectar employs the onlookers to that food source. It afterward chooses another food source close by the one presently in her memory depending on observed information. A new food source is randomly generated by a scout bee to replace the one abandoned by the onlooker bees. This complete search process could be outlined in Fig. 1 as follows [7]:

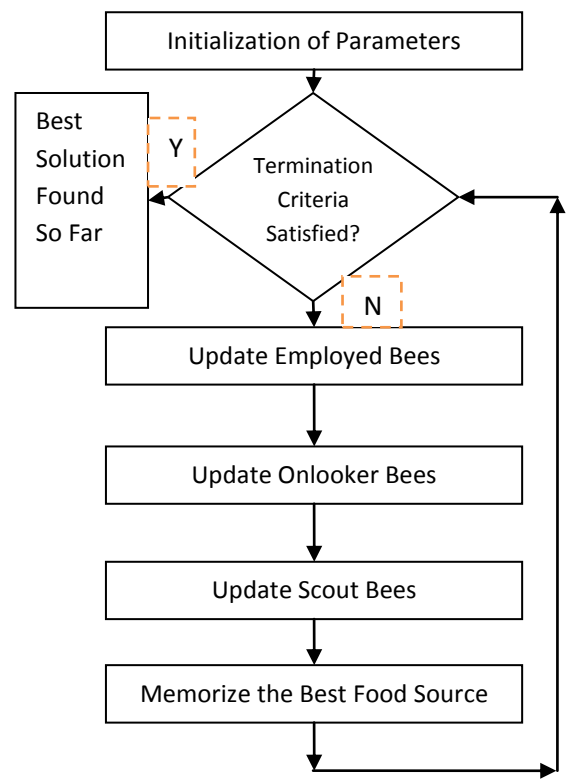

**Fig 1: Phases in Artificial Bee Colony**

### 2.2.1 Initialization of Swarm

The ABC algorithm has three parameters: the number of food sources (population), the number of test after which a food source is treated to be jilted (limit) and the termination criteria (maximum number of cycle). In the original ABC proposed by D. Karaboga [7], the number of food sources is equal to the employed bees or onlooker bees. Initially it consider an evenly dealt swarm of food sources (SN), where each food source $x_i$ (i = 1, 2 ...SN) is a D-dimensional vector. Each food source is generated using following method [8]:

$$x_{ij} = x_{minj} + rand[0,1](x_{maxj} - x_{minj}) \qquad (1)$$

Where

- rand[0,1] is a function that generates an evenly distributed random number in range [0,1].

### 2.2.2 Employed Bee

Employed bees phase update the present solution based on the information of individual experiences and the fitness value of the newly found solution. New food source with higher fitness value replace the existing one. The position update equation for $j^{th}$ dimension of $i^{th}$ candidate during this phase is shown below [8]:

$$V_{ij} = x_{ij} + \varphi_{ij}(x_{ij} - x_{kj}) \qquad (2)$$

Where

- $\varphi_{ij}(x_{ij} - x_{kj})$ is known as step size, k ∈ {1, 2, ..., SN}, j ∈ {1, 2, ...,D} are two randomly chosen indices. k ≠i ensure that step size has some indicative improvement.

### 2.2.3 Onlooker Bee

The number of food sources for onlooker bee is same as the employed. During this phase all employed bee share fitness information of new food sources with onlooker bees. Onlooker bees calculate the selection probability of each food source generated by the employed bee. The best fittest food source is selected by the onlooker. There are number of method for calculation of probability, but it must include fitness. Probability of each food source is decided using its fitness as follow [8]:

$$P_i = \frac{fit_i}{\sum_{i=1}^{SN} fit_i} \qquad (3)$$

### 2.2.4 Scout Bee Phase

If the location of a food source is not updated for a predefined number of cycles, then the food source is assumed to be neglected and scout bees phase is initialized. During this phase the bee associated with the neglected food source converted into scout bee and the food source is replaced by the arbitrarily chosen food source inside the search space. In ABC, the predefined number of cycles is an important control parameter which is called limit for rejection. Now the scout bees replace the abandoned food source with new one using following equation [8].

$$x_{ij} = x_{minj} + rand[0,1](x_{maxj} - x_{minj})$$

$$\forall j = 1,2,......D \qquad (4)$$

Based on the above description, it is clear that in ABC search process there are three main control parameters: the number of food sources SN (same as number of onlooker or employed

bees), the limit and the maximum number of cycles. The algorithm of the ABC is outlined as follow [8]:

| **Algorithm 1:** Artificial Bee Colony Algorithm |
|---|
| Initialize all parameters; |
| Repeat while Termination criteria is not meet |
|       Step 1: Employed bee phase for computing new food sources. |
|       Step 2: Onlooker bees phase for updating location the food sources based on their amount of nectar. |
|       Step 3: Scout bee phase for searching new food sources in place of rejected food sources. |
|       Step 4: Memorize the best food source identified so far. |
| End of while |
| Output: The best solution identified so far. |

## 3. CROSSOVER OPERATOR

The crossover operator is a method for earning genetic information from parents; it combines the features of two parents to form two off-springs, with the probability that good chromosomes may evaluate better ones. The crossover operator is not regularly enforced to all pairs of parent solution the intermediate generation. An incidental choice is made, where the plausibility of crossover being applied depends on probability determined by a crossover rate, known as crossover probability. The crossover operator is most important part in GAs. It combines fraction of good solution to construct new favorable solution. Information involved in one solution mixed with information involved in another solution and the emerging solution will either have good fitness or survive to commutate this information again. If generated two off-springs are identical then crossover operator show strong heritability [25, 26].

### 3.1 Real Coded Crossover Operators

Crossover operators play major role in genetic algorithm which combines the feature of existing solutions and generate new solutions. The optimization problems depend upon the data they used so they are classified in to two categories. One is based on real data set and another one is based on binary or discrete data set. Crossover operator also categorized as binary crossover operators and real coded crossover operators. This paper used the real coded crossover operator for experiment. J C Bansal et al. developed a methodology to share information between two particles using a laplacian operator designed from Laplace probability density function. In this strategy two particles share their positional information in the search space and a new particle is formed. The particle, is known as laplacian particle, replaces the worst performing particle in the swarm. Using this new operator, this paper introduces two algorithms namely Laplace Crossover PSO with inertia weight (LXPSO-W) and Laplace Crossover PSO with constriction factor (LXPSO-C) [27]. A. H. Wright suggests a genetic algorithm that uses real parameter vectors as chromosomes, real parameters as genes, and real numbers as alleles [28].

### 3.1.1 Linear Crossover

Linear crossover [27, 28] is one of the earliest operator in real coded crossover it develops three solutions from two parents and the best two o springs replace parents. Let ( $x_1^{(1,t)}$ $x_2^{(1,t)}$ $x_3^{(1,t)}$ $x_4^{(1,t)}$ ...$x_n^{(1,t)}$ ) and ( $x_1^{(2,t)}$ $x_2^{(2,t)}$ $x_3^{(2,t)}$ $x_4^{(2,t)}$ ...$x_n^{(2,t)}$ ) are two parent solutions of dimension *n* at generation *t*. Linear crossover develops three offspring from these parents as shown in Eq.(5, 6 and 7) and best two offspring being chosen as off-springs.

$$0.5(x_i^{(1,t)}+x_i^{(2,t)}) \tag{5}$$

$$(1.5x_i^{(1,t)}-0.5x_i^{(2,t)}) \tag{6}$$

$$(-0.5x_i^{(1,t)}+1.5x_i^{(2,t)}) \tag{7}$$

Where i = 1, 2, ...., n

# 4. PROPOSED HYBRID CROSSOVER BASED ARTIFICIAL BEE COLONY (CbABC) ALGORITHM

ABC with crossover works in five different phases: first phase is initialization of parameters; second phase is employed bee phase to compute new food sources. Third phase is newly introduces crossover phase. This phase maintains balance between diversification and intensification. Crossover phase diversify the population. Fourth phase is onlooker bee phase to improve solution based on their fitness. Last phase is scout bee phase, this phase search new solutions in place of rejected solutions.

The algorithm of crossover based artificial bee colony (CbABC) algorithm outlined as follow:

---

**Algorithm 2**: Hybrid Crossover Based Artificial Bee Colony (CbABC) Algorithm

---

Initialize all the parameters;
Repeat while Termination criteria is not meet
    Step 1: Employed bee phase for computing new food sources.
    Step 2: Crossover phase to increase the quality of solution.
    Step 3: Onlooker bees phase for updating the location of food sources based on their nectar amounts.
    Step 4: Scout bee phase for searching new food sources in place of rejected food sources.
    Step 5: Memorize the best food source identified so far.
End of while loop
    Output: Best solution identified so far.

---

The first step consists of the evaluation of the population using the Artificial Bee Colony. Initial populations generated by ABC are used by employed bees. After this crossover operators are applied. If crossover criteria or probability fulfilled than two random parents are taken to perform crossover operation on them. After crossover operation new off-springs are developed. Worst parent replaced by best developed offspring if its fitness is better than the worst parent. Crossover operator is applied to two arbitrarily selected parents from current population. Two offspring developed from crossover and worst parent is replaced by best offspring, other parent remains same. The complete process repeats itself until the maximum numbers of cycles are completed. The phases of modified ABC algorithm are outlined in Fig. 2.

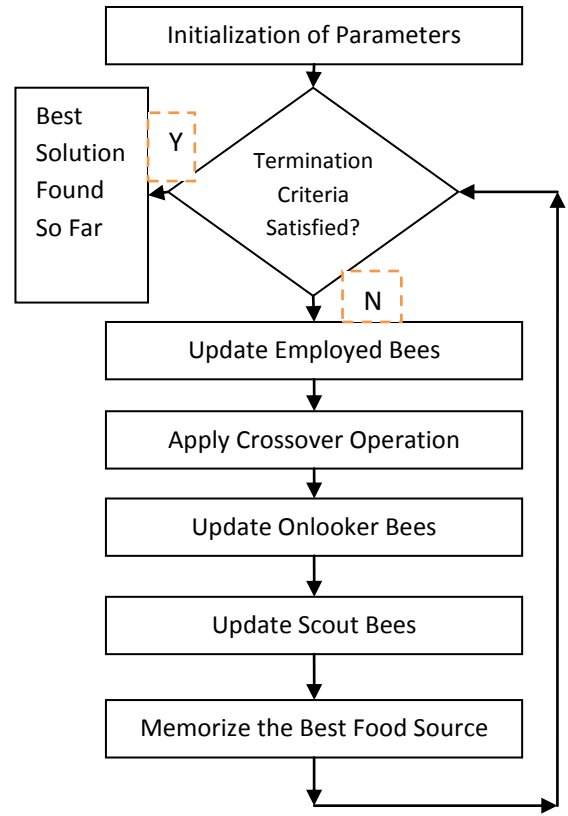

**Fig 2: Phases in Hybrid Crossover based Artificial Bee Colony (CbABC) Algorithm**

In this paper crossover applied in each iteration after employed bee phase. The detailed pseudo code of proposed crossover based ABC algorithm is outlined in algorithm 3.

---

**Algorithm 3**: Detailed Hybrid Crossover Based ABC (CbABC) Algorithm

---

Initialization phase
For $i = 0$ to the maximum no. of food source size do
 For $j = 0$ to the problem dimension
  Initialize all food sources arbitrarily using equation (1)
 End for $j$
Evaluate fitness for all food sources
End for $i$
R e p e a t
Employed Bee Phase
For $i=0$ to maximum no. of employed bees
 For $j=0$ to maximum problem dimension
  Generate new candidate solution using equation (2)
 End for $j$
 Compute fitness values for all new generated candidate solution
 If candidate solution is better than the old solution then replace old solution with candidate solution
End for $i$
Crossover phase
 If crossover criteria satisfies
 For $i = 0$ to the maximum no. of food source
  Select two random individuals from the current population for crossover operation.
  Apply crossover operation.
  New off-springs generated from parents as a result of crossover. Replace the worst parent with the best new offspring if it is better.
  End for *i*
Onlooker Bee phase
  For *i = 0* to the maximum no. of onlooker bees
   For *j=0* to maximum dimension
   Generate new candidate solution
   End for *j*
  Calculate the fitness for new generated candidate solution
  If candidate solution has the better fitness values then replace old solution
  End for *i*
Scout Bee phase
  If any food source exhausted
  Initialize randomly exhausted food source until maximum cycle no.

# 5. EXPERIMENTAL RESULTS

## 5.1 Test problems under consideration
Artificial Bee Colony algorithm with linear crossover applied to the four benchmark functions for whether it gives better result or not at different probability. Benchmark functions taken in this paper are of different characteristics like uni-model (Sphere, Rosenbrock) or multi-model (Griewank, Rastrigin) and separable (Sphere, Rastrigin) or non-separable (Griewank, Rosenbrock). In order to analyze the performance of CbABC, 4 different global optimization problems ($f_1$ to $f_4$) are selected (listed in Table 1). These are continuous optimization problems and have different degrees of complexity and multimodality. Test problems $f_1$ –$f_4$ are taken from [14].

### Table 1. Test problems

| Test Problem | Objective function | Characteristic | Ranges |
|---|---|---|---|
| Sphere | $f_1(\vec{x}) = \sum_{i=1}^{D} x_i^2$ | Uni-model, Separable | $-100 \leq x_i \leq 100$ |
| Griewank | $f_2(\vec{x}) = \frac{1}{4000}(\sum_{i=1}^{D}(x_i^2)) - (\prod_{i=1}^{D} \cos(\frac{x_i}{\sqrt{i}})) + 1$ | Multi-model, Non-separable | $-600 \leq x_i \leq 600$ |
| Rastrigin | $f_3(\vec{x}) = \sum_{i=1}^{D}(x_i^2 - 10\cos(2\pi x_i) + 10)$ | Multi-model, Separable | $-5.12 \leq x_i \leq 5.12$ |
| Rosenbrock | $f_4(\vec{x}) = \sum_{i=1}^{D} 100(x_i^2 - x_{i+1})^2 + (1 - x_i)^2$ | Uni-model, Non-separable | $-30 \leq x_i \leq 30$ |

## 5.2 Experimental Setup
To prove the efficiency of CbABC, it is compared with original ABC over considered four problems, following experimental setting is adopted:
  – The size of colony= Population size SN =20
  – Number of Employed bee = Number of Onlooker bee =SN/2=10

  – The maximum number of cycles for foraging MCN =2000
  – Number of repetition of experiment =Runtime =30
  – Dimension of Problem space D =30
  – Limit =100 ,A food source which could not be improved through "limit" trial is abandoned by its employed bee
  – 4 Benchmark functions are used and each has minimum value =0
  – The mean function values of the best solutions found by the algorithms for different dimensions have been recorded.

## 5.3 Result Comparison
Results with experimental setting discussed above are outlined in table 2. In table 2 mean objective (MO) and average evaluation (AE) are reported. In this paper original ABC compared with linear crossover (LX). Table 2 shows that most of the time Crossover based ABC outperformed original ABC in terms of performance.

### Table 2 Comparison of the results of Test problems

| Test Function | Measure | ABC | Pr | LX |
|---|---|---|---|---|
| $f_1$ | MO | 4.589e-02 | 0.1 | 4.596e-02 |
| | | | 0.2 | 4.612e-02 |
| | | | 0.3 | 4.565e-02 |
| | AE | 2167 | 0.1 | 2296 |
| | | | 0.2 | 2248 |
| | | | 0.3 | 2386 |
| $f_2$ | MO | 2.417e-06 | 0.1 | 2.548e-06 |
| | | | 0.2 | 2.561e-06 |
| | | | 0.3 | 2.428e-06 |
| | AE | 5137 | 0.1 | 5190 |
| | | | 0.2 | 5307 |
| | | | 0.3 | 5567 |
| $f_3$ | MO | 276.3 | 0.1 | 271.6 |
| | | | 0.2 | 277.2 |
| | | | 0.3 | 279.7 |
| | AE | 20000 | 0.1 | 20000 |
| | | | 0.2 | 20000 |
| | | | 0.3 | 20000 |
| $f_4$ | MO | | 0.1 | 9.546 |

| | 9.816 | 0.2 | 9.582 |
| | | 0.3 | 9.715 |
| | | 0.1 | 20000 |
| AE | | 0.2 | 20000 |
| | 20000 | 0.3 | 20000 |

Obtained results indicate that ABC With appropriate crossover and appropriate crossover probability may provide far better results than original ABC. However there is no fixed value of crossover probability for which crossover operator can improve the results generated by ABC. However, from the applications point of view a range from 0.1 to 0.3 is the most suitable for crossover probability as shown by tables, crossover operator when applied to basic ABC algorithm improves the results in terms of performance and accuracy. CbABC algorithm improves the results of TSP as compared to ABC algorithm.

Table 2 lists the comparison results of these operators for 4 benchmark functions of 3 different probabilities for ABC algorithm. Each of experiments was repeated 30 times, and the mean objective (MO) and average evaluation (AE) are listed in the tables, of which the bold mean values are the best results among the listed methods. From the above table it can be observed that with the probability increasing, the mean objective values of most of the benchmark functions are decreased. For almost all functions most of the time Crossover based ABC outperformed original ABC in terms of performance..

# 6. APPLICATION OF PROPOSED ALGORITHM

To measure the robustness of the proposed strategy, a real world optimization problem, namely travelling salesman problem considered. The Traveling Salesman Problem (TSP) is one of the hardest problems for which no efficient method is known. Assuming a travelling salesman has to travel a given number of cities, starting and ending at the same city. This tour, which represents the length of the travelled path, is the formulation of TSP. As the number of cities increases, the determination of the optimal tour (also known as Hamiltonian tour), becomes very complex. A TSP decision problem is generally classified as NP-Complete problem.

The TSP can be defined on a complete undirected graph $G = (V, E)$ if it is symmetric or on a directed graph $G = (V, A)$ if it is asymmetric. Here V, E and A are set of vertices, edges and arcs respectively. The set $V = \{1, \ldots, n\}$ is the vertex set, $E = \{(i, j) : i, j \in V, i < j\}$ is an edge set and $A = \{(i, j) : i, j \in V, i \neq j\}$ is an arc set. A cost matrix $C = (c_{ij})$ is defined either on E or on A. The cost matrix satisfies the triangle inequality whenever $c_{ij} \leq c_{ik} + c_{kj}$, for all values of $i, j, k$. specially, this is the case of planar problems for which the vertices are points $P_i = (x_i, y_i)$ in the plane, and $c_{ij} = \sqrt{(x_i - x_j)^2 + (y_i - y_j)^2}$ is known as the Euclidean distance. The triangle inequality is also satisfied if $c_{ij}$ is the length of a shortest path from $i$ to $j$ on graph G.

Experimental setup for finding solution of TSP is same as it was for standard benchmark function except that now it is tested for different number of cycles. Number of cycles varies from 500 to 3000. CbABC use linear crossover operator for solving travelling salesman problem.

## 6.1 Experimental Results

The CbABC using linear crossover operator outperformed ordinary ABC. Results are taken for dimension 10, 20 and 30 with varying maximum cycle numbers (from 500 to 3000). Table 3 summarizes these results. Simple ABC algorithm degrades performance when maximum cycle number exceeds 1500 but CbABC continuously keeps improving results.

**Table 3 Comparison of the results for Travelling Salesman problem**

| Algorithm | Dimension | Maximum Number of Cycles | | | | | |
|---|---|---|---|---|---|---|---|
| | | 500 | 1000 | 1500 | 2000 | 2500 | 3000 |
| ABC | 10 | 229.83 | 220.40 | 213.07 | 216.27 | 212.53 | 214.14 |
| | 20 | 289.83 | 271.93 | 264.90 | 261.73 | 257.54 | 252.83 |
| | 30 | 394.40 | 365.83 | 362.26 | 352.46 | 346.91 | 341.23 |
| CbABC | 10 | 138.30 | 137.40 | 132.90 | 129.50 | 131.15 | 129.05 |
| | 20 | 152.65 | 128.20 | 90.70 | 89.75 | 82.70 | 75.80 |
| | 30 | 81.30 | 67.15 | 59.00 | 57.90 | 54.35 | 48.65 |

The proposed algorithm compared on the basis of dimension and maximum number of cycles. The CbABC algorithm improves the results of TSP as compared to ABC algorithm. CbABC improves result, when dimension increased. Fig.4-6 shows that how CbABC outperformed simple ABC when applied for TSP.

# 7. CONCLUSION

This work presents a novel hybrid optimization algorithm named CbABC. This algorithm integrates crossover operator from GA with ABC. Linear crossover operator is applied to the Artificial Bee Colony algorithm. The worst parent replaced by an off-spring generated by two randomly selected parents in each cycle, if fitness of generated off-spring is better than the worst parent from the current swarm. The experiments are performed on a set of four well known benchmark functions. Crossover probability range [0.1 – 0.3] considered for experiments. It is observed that linear crossover operator with ABC is better choice for continuous optimization. Performance of proposed algorithm refined with increment in crossover probability. It also shows that CbABC algorithm improves the TSP results in comparison to original ABC algorithm. Numerical results show that the proposed method is superior to ABC approach. Proposed algorithm has the ability to get out of a local minimum and can be efficiently used for separable, multivariable, multimodal function optimization.

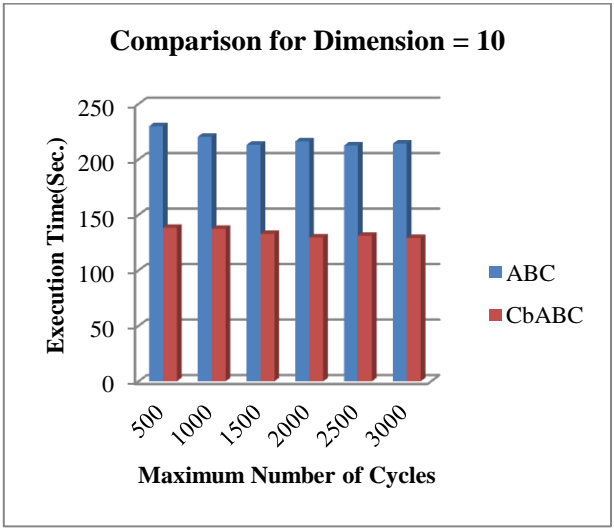

**Fig 3: Comparison of ABC and CbABC for TSP at D = 10**

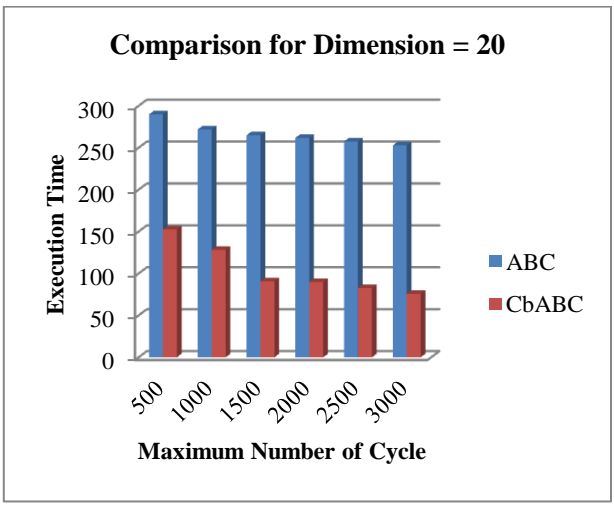

**Fig 4: Comparison of ABC and CbABC for TSP at D = 20**

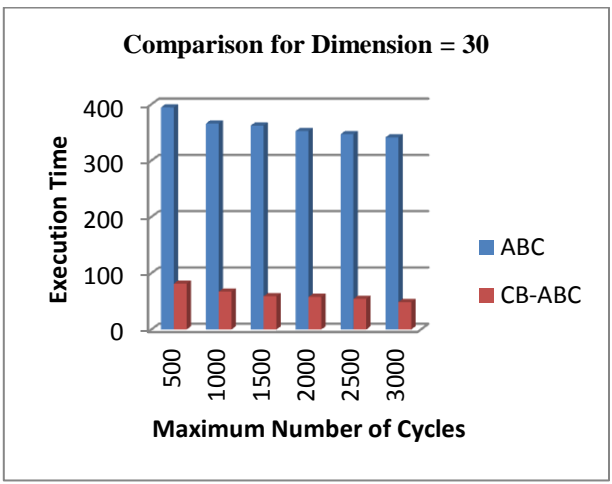

**Fig 5: Comparison of ABC and CbABC for TSP at D = 30**